\documentclass[10pt,twocolumn,letterpaper]{article}

\usepackage{wacv}
\makeatletter
\@namedef{ver@everyshi.sty}{}
\makeatother
\usepackage{tikz}
\usepackage{times}
\usepackage{epsfig}
\usepackage{graphicx}
\usepackage{amsmath}
\usepackage{amssymb}
\usepackage{booktabs}

\usepackage{tikz}
\usepackage{comment}
\usepackage{amsmath,amssymb} 
\usepackage{color}

\usepackage[accsupp]{axessibility}  

\usepackage{multirow}
\usepackage{threeparttable}
\usepackage{float}
\usepackage{booktabs}
\usepackage{pifont}
\usepackage{gensymb}
\usepackage[normalem]{ulem}

\usepackage{wrapfig}

\definecolor{lowresfeat}{HTML}{6C8EBF}
\definecolor{feedback}{HTML}{3700CC}
\definecolor{shuffleconv}{HTML}{0050EF}
\definecolor{fa}{HTML}{FFF2CC}
\definecolor{ob}{HTML}{DAE8FC}
\definecolor{warpedfeat}{HTML}{A50040}
\definecolor{flowfield}{HTML}{2D7600}
\definecolor{dcnprop}{HTML}{D79B00}
\definecolor{innerfeat}{HTML}{9673A6}
\definecolor{fovea}{HTML}{6F0000}
\definecolor{dsv}{HTML}{9673A6}

\def\wacvPaperID{643} 

\wacvalgorithmstrack   

\wacvfinalcopy 

\ifwacvfinal
\usepackage[breaklinks=true,bookmarks=false]{hyperref}
\else
\usepackage[pagebackref=true,breaklinks=true,colorlinks,bookmarks=false]{hyperref}
\fi

\begin{document}

\title{Cross-Resolution Flow Propagation for Foveated Video Super-Resolution}

\author{Eugene Lee\quad Lien-Feng Hsu \quad Evan Chen \quad Chen-Yi Lee\\
National Yang Ming Chiao Tung University\\
Hsinchu, Taiwan\\
{\tt\small  \{eugene.ee06g,lienfeng.ee09g,evanchen.ee06\}@nctu.edu.tw, cylee@si2lab.org}
\and
}

\maketitle
\thispagestyle{empty}


\begin{abstract}
The demand of high-resolution video contents has grown over the years. However, the delivery of high-resolution video is constrained by either computational resources required for rendering or network bandwidth for remote transmission. To remedy this limitation, we leverage the eye trackers found alongside existing augmented and virtual reality headsets. We propose the application of video super-resolution (VSR) technique to fuse low-resolution context with regional high-resolution context for resource-constrained consumption of high-resolution content without perceivable drop in quality.
Eye trackers provide us the gaze direction of a user, aiding us in the extraction of the regional high-resolution context.
As only pixels that falls within the gaze region can be resolved by the human eye, a large amount of the delivered content is redundant as we can't perceive the difference in quality of the region beyond the observed region.
To generate a visually pleasing frame from the fusion of high-resolution region and low-resolution region, we study the capability of a deep neural network of transferring the context of the observed region to other regions (low-resolution) of the current and future frames.
We label this task a Foveated Video Super-Resolution (FVSR), as we need to super-resolve the low-resolution regions of current and future frames through the fusion of pixels from the gaze region.
We propose Cross-Resolution Flow Propagation (CRFP) for FVSR.
We train and evaluate CRFP on REDS dataset on the task of $8\times$ FVSR, i.e.\ a combination of $8\times$ VSR and the fusion of foveated region.
Departing from the conventional evaluation of per frame quality using SSIM or PSNR, we propose the evaluation of past foveated region, measuring the capability of a model to leverage the noise present in eye trackers during FVSR.
Code is made available at \url{https://github.com/eugenelet/CRFP}.
\end{abstract}


\section{Introduction}
The impact of video super-resolution (VSR) in our daily life has become more prominent in the recent years as high quality contents can be delivered while its lower quality counterpart is rendered or stored, saving either computational or storage resources.
The application of deep neural networks to the task of rendering high-resolution frames using its low-resolution sampled counterpart has brought forward substantial improvements that enables technologies like Deep Learning Super-Sampling (DLSS) \cite{edelsten2019truly} and DeepFovea \cite{kaplanyan2019deepfovea,xiao2020neural}.
They deliver high quality content on a computationally-constrained platform.
While existing VSR techniques are implemented on a pixel level and are able to reconstruct video content to a point that is visually pleasing, certain context that are of high-quality that is meant to be delivered might not be fully reconstructed, restricting high frequency context from being delivered, e.g.\ texts and fine textures.
Results of VSR techniques are visually acceptable up to $4\times$ VSR, while frames generated using $8\times$ VSR have distinctive flaws that affects the overall viewing experience.
We argue that VSR methods while being useful for delivering general contexts, it should open up the possibility of fusing super-resolved frames with regional high-resolution (HR) context(s), e.g.\ HR patches, that are crucial for the understanding of the gist of the delivered content.

With the increase in adoption of augmented and virtual reality (AR/VR) devices \cite{fernandez2017augmented,wei2019research,hassan2018augmented}, the demand for high-resolution content will show similar spike.
As more pixels are required for the immersive experience for AR/VR, developers are searching for effective ways to reduce the computational cost of rendering frames for AR/VR.
A feasible approach is to include an eye tracker in the AR/VR headset to estimate the gaze direction of the user \cite{clay2019eye}.
Frames are rendered based on the gaze direction of the user \cite{stengel2016adaptive,kaplanyan2019deepfovea}, resulting in huge reduction in computational cost.
Our work leverages the eye tracker of such devices for the task of Foveated Video Super-Resolution (FVSR).
FVSR is useful if we are to transfer HR content to be viewed in real-time, especially to AR/VR devices.
Transferring the HR frames at its full resolution might not be feasible at a bandwith-constrained environment.
For FVSR, only the pixels that fall in gaze region are transmitted in HR while the rest are transmitted in low-resolution (LR).
This results in huge savings in bandwidth as it is empirically shown that the human eye is only able to perceive and resolve around $\sim$1\% of pixels in a frame \cite{wang2021focas,guenter2012foveated}.
The main challenge of FVSR is the transfer of context from the HR to the LR region, to prevent abrupt transition in visual quality.


As prior works of VSR don't consider the fusion of LR and HR context, cross-temporal operation or alignment is performed on the feature maps of lowest spatial resolution, i.e.\ during the propagation and aggregation stages \cite{chan2021basicvsr,chan2021basicvsr++}.
The temporally-aggregated low spatial resolution feature maps are then upsampled using a sequence of upsampling filters to reconstruct the HR frame.
To incorporate regional HR context(s) into the super-resolution pipeline, we need to have precise spatial locality for the placement of the HR region.
To do so, we propose a Cross-Resolution Flow Propagation (CRFP) framework that follow existing VSR framework which sequentially performs propagation, alignment, aggregation and upsampling.
To aggregate the foveated context into the super-resolution pipeline, few modifications are made.
Foveated region are fed to the \textit{Feature Aggregator} (FA) using a feedback mechanism.
Multiple FAs are placed at the features of lowest resolution and a single FA is placed before an output block with features having the targeted resolution.
As the spatial resolution of the feature maps of the final stage (after upsampling stages) matches the spatial resolution of the HR region, i.e.\ matching coordinates, the spatial fusion of both features can be precise.
Leveraging VSR techniques for the construction of CRFP, we show promising results for FVSR that are not achievable by existing VSR techniques.


The closest work to our proposed research direction of FVSR is DeepFovea \cite{kaplanyan2019deepfovea}. DeepFovea performs video inpainting given a sequence of sparse frames.
As FVSR is a novel task, the only comparison we make is with the architecture we bootstrap CRFP on, BasicVSR++ \cite{chan2021basicvsr++} (SoTA in VSR), modified for the task of FVSR.
Our contributions are summarized as follows:
\begin{enumerate}
    \item We propose a new task of Foveated Video Super-Resolution (FVSR). FVSR requires an eye tracker to work and is applicable to the growing adoption of AR/VR devices for the streaming of HR video.
    \item We propose a Cross-Resolution Flow Propagation (CRFP) technique for FVSR, demonstrating convincing results for FVSR.
    \item To quantitatively measure the performance of FVSR, we propose the evaluation of Past Foveated Region using PSNR and SSIM to better evaluate the capability of a model to retain contexts from previous frames.
\end{enumerate}

\section{Background and Related Work}  
\paragraph{Visual Perception of Foveated Video.}
As video contents are designed to be consumed by the human eye, we can exploit how visual signal is encoded for processing at the visual system to compress our data source without inducing perceptible loss in visual quality. Curcio \etal\ \cite{curcio1990human} shows that there's a rapid decrease in the number of photoreceptors in the eye from the fovea to the periphery, also known as eccentricity.
Despite the loss in spatial resolution, Rovamo \etal\ \cite{rovamo1984temporal} shows that temporal sensitivity remains static spatially, requiring the displayed video to have smooth transition across frames. The perception of spatial detail at a certain spatial frequency (visual acuity) is limited by the density of the midget ganglion cells that provide the pathway out of the eye \cite{kelly1984retinal,robson1966spatial}. Dacey and Patersen \cite{dacey1992dendritic} show that there's an order of $30\times$ reduction in cell density from the fovea to periphery (0\degree - 40\degree), giving us a hint on the size of the foveated region to be cropped from the HR image.
The central 5.2\degree region of the retina has high sensitivity, covering only 0.8\% of total pixels on a regular display \cite{wang2021focas,guenter2012foveated}.
This finding points us to the choice of cropping $\sim$1\% of the total pixels as the foveated region in our experiments.

Studies in \cite{thibos1996characterization,wang1996undersampling} shows that in peripheral regions, mismatch between optical, retinal and final neural sampling resolutions leads to aliasing in our peripheral vision. \textit{Aliasing zone} is the gap between the detection and resolution thresholds \cite{patney2016towards}. Context between the detection and resolution threshold are details can be detected but not resolved whereas context within the resolution threshold can be clearly resolved and detected. The role of FVSR is to attempt to reconstruct the context of the targeted frame such that the quality of the reconstructed pixels within the aliasing zone is visually pleasing. This is measured by the outskirt of foveated region in the experimental section.
Naive downsampling of video with eccentricity will introduce aliasing and jitter effect when viewed. Guenter \etal\ \cite{guenter2012foveated} progressively compute three gaze-centered concentric rings to address this problem. Stengel \etal\ \cite{stengel2016adaptive} propose to perform sparse rendering in the periphery with either stochastic sampling and inpainting. Temporal models from VSR are referred for the design of models for FVSR \cite{chan2021basicvsr++,wang2019edvr}.

\paragraph{Single Image Super-Resolution.}
Early work on super-resolution processes each frame seperately. SRCNN is a simple 3-layer super-resolution convolutional neural network proposed by Dong \etal\ \cite{dong2014learning}. Kim \etal\ \cite{kim2016accurate} explores a deeper architecture, VDSR, a 20-layer deep network with residual connections. ResNet \cite{he2016deep} and generative adversarial networks \cite{goodfellow2014generative} is adopted by Ledig \etal\ \cite{ledig2017photo} in SRGAN and Sajjadi \etal\ \cite{sajjadi2017enhancenet} in EnhanceNet to generate high-frequency detail. Tai \etal\ \cite{tai2017image} propose DRRN that uses recursive residual blocks. Tong \etal\ propose SRDenseNet \cite{tong2017image} which uses DenseNet \cite{huang2017densely} as its backbone. Pan \etal\ \cite{pan2018learning} propose DualCNN that uses two branches to reconstruct structure and detail components of an image.

\paragraph{Video Super-Resolution.}
To exploit temporal information across frames in a video, temporal alignment or motion compensation is used either explicitly or implicitly. Explicit VSR makes use of information from neighboring frames through motion estimation and compensation. Earlier work for motion estimation is based on optical flow, e.g.\ Liao \etal\ \cite{liao2015video} uses optical flow methods \cite{xu2011motion} along with a deep draft-ensemble network to reconstruct the HR frame. Kappler \etal\ \cite{kappeler2016video} predicts HR frame by taking interpolated flow-wrapped frames as inputs to a CNN. VESPCN \cite{caballero2017real} is the first VSR work that jointly trains flow estimation and spatio-temporal networks. SPMC \cite{tao2017detail} uses an optical flow network to compute LR motion field to generate sub-pixel information to achieve sub-pixel motion compensation. TOFlow \cite{xue2019video} shows that task-oriented motion cues achieves better VSR results than fixed flow algorithms. RBPN \cite{haris2019recurrent} propose a recurrent encoder-decoder module to exploit inter-frame motion that is estimated explicitly. EDVR \cite{wang2019edvr} uses a deformable convolution module to align multiple frames to a reference frame in feature space and uses a temporal and spatial attention module for fusion. Jo \etal\ \cite{jo2018deep} are the first to propose the use of dynamic upsampling filters (DUF) for VSR. To consider neighboring frames for implicit VSR, a common approach is by using a sliding window, i.e\ the concatenation of images within a fixed window length \cite{xue2019video,jo2018deep,haris2019recurrent,wang2019edvr,isobe2020video2}. Most sliding window methods are symmetric, i.e.\ past and future frames are considered for the reconstruction of the targeted frame (non-causal), making them unsuitable for streaming applications. Recurrent methods \cite{sajjadi2018frame,fuoli2019efficient,isobe2020video} passes information from previous frames through a hidden representation.
Our work is based on the idea of flow-guided deformable alignment from BasicVSR++ \cite{chan2021basicvsr++}.
Such idea has been earlier studied by several works \cite{jia2017controllable,sun2018pwc}.

\section{Methods}

\begin{figure*}[t]
    \centering
    \includegraphics[width=\textwidth]{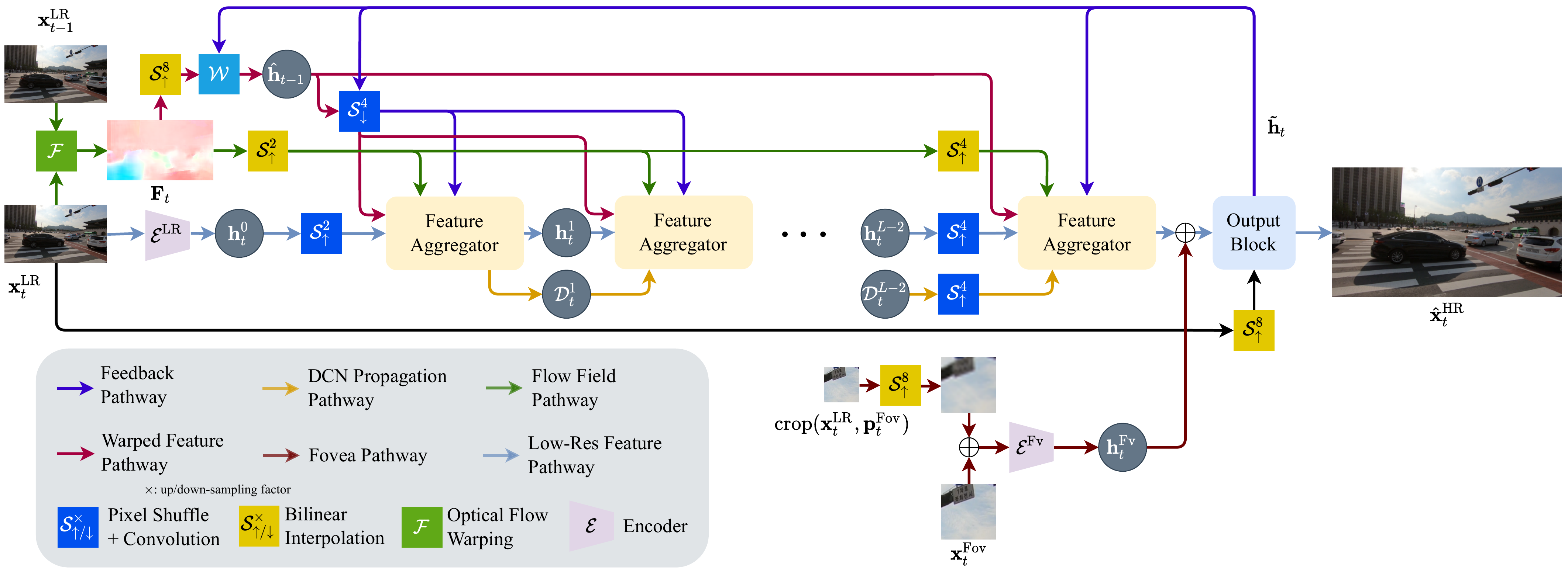}
    \caption{Overview of Cross-Resolution Flow Propagation for $8\times$ FVSR. The core building blocks are \textcolor{fa}{Feature Aggregator (FA)} and \textcolor{ob}{Output Block (OB)}. Foveated region is regionally aggregated at the \textcolor{ob}{OB}. The \textcolor{ob}{OB} has two outputs, the super-resolved frame $\hat{\mathbf{x}}_t^{\mathrm{HR}}$ and the fovea-aggregated feature $\tilde{\mathbf{h}}_t$ that is fed to earlier stages of the network through the \textcolor{feedback}{Feedback Pathway}. Features at the lower spatial resolution encoded by $\mathcal{E}^{\mathrm{LR}}$ and features of the highest spatial resolution right after the upsampling (\textcolor{shuffleconv}{Pixel Shuffle + Convolution}) block $\mathcal{S}_{\mathrm{s}\uparrow}^4$ that passes through the \textcolor{lowresfeat}{Low-Res Feature Pathway}, are aggregated with features from other pathways (\textcolor{feedback}{Feedback}, \textcolor{dcnprop}{DCN Propagation}, \textcolor{flowfield}{Flow Field}, \textcolor{warpedfeat}{Warped Feature}) through the \textcolor{fa}{FA}. Repeated adoption of \textcolor{fa}{FA} sharing the same input and output connection pattern is abbreviated as ``$\cdot\cdot\cdot$'' in the illustration. We use the same downsampling (\textcolor{shuffleconv}{Pixel Shuffle + Convolution}) block $\mathcal{S}_{\mathrm{s}\downarrow}^4$ (tied-weights) to encode features from the \textcolor{feedback}{Feedback Pathway} and \textcolor{warpedfeat}{Warped Feature Pathway} for the \textcolor{fa}{FA}. }
    \label{fig:crfp_illus}
\end{figure*}


We propose Cross-Resolution Flow Propagation (CRFP), a novel framework for FVSR.
CRFP is able to aggregate context from gaze region that is of high resolution (HR) to the low resolution (LR) counterpart.
To provide high fidelity video stream, HR context of previous frames should be captured and retained by the framework such that future frames can be better super-resolved using the retained context.
This design works in tandem with the nature of eye tracking devices.
The gaze coordinate predicted by eye tracking devices is usually corrupted by additive Gaussian noise, $\mathbf{p}_t^{\mathrm{Fov}}\in \mathcal{N}(\mu_t^{\mathrm{Fov}},\sigma_t^{\mathrm{T}})$, having the predicted gaze coordinate oscillating around the actual gaze direction $\mu_t^{\mathrm{Fov}}$ under a Gaussian noise of the eye tracker of standard deviation $\sigma_t^{\mathrm{T}}$.
This is analogous to the application of super-resolution techniques to handheld cameras \cite{wronski2019handheld,lecouat2021lucas,bhat2021deep}, where the natural hand tremor is exploited during the reconstruction of the original frame.
The better a model is at capturing and retaining HR context from past foveated region, the better it can exploit the prediction noise from the eye tracker.
In Section \ref{sec:crfp} we discuss the pathways in our architecture that contributes to the retention of context from past foveated region.
In Section \ref{sec:aggregator} we provide in-depth description of the Feature Aggregator.
In Section \ref{sec:output} we show how context from the foveated region is aggregated into the feedback and frame generation pipeline.
We illustrate an overview of CRFP in Figure \ref{fig:crfp_illus}.

\subsection{Cross-Resolution Flow Propagation}\label{sec:crfp}
To adopt the HR context for the super-resolution of LR context corresponding to current and future frames, an architecture that focuses on cross-resolution propagation of context is required.
CRFP is proposed to handle this problem, introducing two core building blocks for cross-resolution context aggregation, namely the \textcolor{fa}{\textit{Feature Aggregator} (FA)} and \textcolor{ob}{\textit{Output Block} (OB)}.
These building blocks are connected by several information pathways, i.e.\ \textcolor{feedback}{Feedback Pathway}, \textcolor{dcnprop}{DCN Propagation Pathway}, \textcolor{flowfield}{Flow Field Pathway}, \textcolor{warpedfeat}{Warped Feature Pathway}, \textcolor{fovea}{Fovea Pathway} and \textcolor{lowresfeat}{Low-Res Feature Pathway}, each playing different roles in the aggregation process.
The goal of FVSR is to super-resolve the LR frame at timestep $t$, $\mathbf{x}_t^{\mathrm{LR}}$, while considering an additional foveated region of HR, $\mathbf{x}_t^{\mathrm{Fov}}$.
Without any external factors, the LR frame $\mathbf{x}_t^{\mathrm{LR}}$ propagates through the Low-Res Feature Pathway.
$\mathbf{x}_t^{\mathrm{LR}}$ is first encoded by an encoder $\mathcal{E}^{\mathrm{LR}}$ followed by a pixel shuffle + convolution block $\mathcal{S}_{\mathrm{s}\uparrow}^2$ for $2\times$ up-sampling, giving us $\mathbf{h}_t^{0}$,
\begin{equation}
    \mathbf{h}_t^0 = \mathcal{S}_{\mathrm{s}\uparrow}^2 \left( \mathcal{E}^{\mathrm{LR}} (\mathbf{x}_t^{\mathrm{LR}}) \right).
\end{equation}
The encoded features are then fed to several FA blocks along the Low-Res Feature Pathway to be aggregated with information from other pathways,
\begin{align}
    \{\textcolor{lowresfeat}{\mathbf{h}_t^{l+1}}, \textcolor{dcnprop}{\mathcal{D}_t^{l+1}} \}=& \nonumber \\
    \textcolor{fa}{\mathrm{FA}^l}(\mathcal{S}_{\mathrm{s}\downarrow}^4(\textcolor{lowresfeat}{\mathbf{h}_t^l}); \textcolor{warpedfeat}{\hat{\mathbf{h}}_{t-1}}, \textcolor{flowfield}{\mathbf{F}_t},& \mathcal{S}_{\mathrm{s}\downarrow}^4(\textcolor{feedback}{\tilde{\mathbf{h}}_{t-1}}), \textcolor{dcnprop}{\mathcal{D}_t^l}, \mathcal{W}(\textcolor{innerfeat}{\mathbf{z}_t^l}; \textcolor{flowfield}{\mathbf{F}_t)} ), \nonumber\\
    & \qquad  \qquad  l=0,...,L-1.
\end{align}
In our design, we have $L=4$ where the first three FAs are placed at the feature of the lowest spatial resolution and one FA placed after the up-sampling stage $\mathcal{S}_{\mathrm{s}\uparrow}^4$.
The motivation of such placement is to enable the aggregation of information at different spatial resolution while keeping computational cost low. 
Placing FA after the up-sampling stage would increase the computational cost quadratically in accordance to the up-sampling rate but the aggregation of HR context is more precise since it's closer to the coordinate system of the pixel space.
The output of the final FA $\mathbf{h}_t^{L-1}$ is concatenated ($\oplus$) with the encoded foveated region $\mathbf{h}_t^{\mathrm{Fv}}$ as input for the OB to render the super-resolved frame $\hat{\mathbf{x}}_t^{\mathrm{HR}}$ and to estimate the fovea-fused feature $\tilde{\mathbf{h}}_t$ to be propagated to earlier layers through the Feedback Pathway for aggregation with features of future frames,
\begin{equation}
    \left\{\hat{\mathbf{x}}_t^{\mathrm{HR}}, \textcolor{feedback}{\tilde{\mathbf{h}}_t }\right\} = \textcolor{ob}{\mathrm{OB}}\left(\textcolor{lowresfeat}{\mathbf{h}_t^{L-1}}\oplus \textcolor{fovea}{\mathbf{h}_t^{\mathrm{Fv}}}, \mathcal{S}_{i\uparrow}^8(\mathbf{x}_t^{\mathrm{LR}} )\right).
\end{equation}
Note that OB also takes in the bilinearly-upsampled LR frame $\mathcal{S}_{\mathrm{i}\uparrow}^8(\mathbf{x}_t)^{\mathrm{LR}}$ with details deferred to Section \ref{sec:output}.
$\mathbf{h}_t^{\mathrm{Fv}}$ originates from the Fovea Branch,
\begin{equation}
    \textcolor{fovea}{\mathbf{h}_t^{\mathrm{Fv}}} = \mathcal{E}^{\mathrm{Fv}}\left(\textcolor{fovea}{\mathbf{x}_t^{\mathrm{Fv}}} \oplus \mathrm{crop}( \mathcal{S}_{\mathrm{i}\uparrow}^{8}(\mathbf{x}_t^{\mathrm{LR}}), \textcolor{fovea}{\mathbf{p}_t^{\mathrm{Fv}}})\right).
\end{equation}
$\mathbf{h}_t^{\mathrm{Fv}}$ is the result of a fovea encoder $\mathcal{E}^{\mathrm{Fv}}$ applied to the concatenation of the HR foveated region $\mathbf{x}_t^{\mathrm{Fv}}$ and the $8\times$ bilinearly up-sampled LR frame $\mathbf{x}_t^{\mathrm{LR}}$ cropped ($\mathrm{crop}$) using gaze coordinate $\mathbf{p}_t^{\mathrm{Fv}}$.
The fovea-fused feature $\tilde{\mathbf{h}}_{t-1}$ from the previous time-step is down-sampled using a pixel shuffle + convolution block $\mathcal{S}_{\mathrm{s}\downarrow}^4$ to match the spatial resolution of the features of earlier layers in the FA.
The same down-sampling block is utilized to down-sample the warped version of $\tilde{\mathbf{h}}_{t-1}$,
\begin{equation}
    \textcolor{warpedfeat}{\hat{\mathbf{h}}_{t-1}} = \mathcal{W}\left(\textcolor{feedback}{\tilde{\mathbf{h}}_{t-1}}, \mathcal{S}_{\mathrm{i}\uparrow}^8(\textcolor{flowfield}{\mathbf{F}_t})\right).
\end{equation}
$\mathcal{W}$ is the warping operator that warps an input image using optical flow $\mathbf{F}_t$.
$\mathbf{F}_t$ is bilinearly up-sampled using $\mathcal{S}_{\mathrm{i}\uparrow}^8(\cdot)$ to match the size of $\tilde{\mathbf{h}}_{t-1}$.
$\mathbf{F}_t$ is estimated using an optical flow estimator $\mathcal{F}$ based on frames from time-steps $t$ and $t-1$,
\begin{equation}
    \textcolor{flowfield}{\mathbf{F}_t} = \mathcal{F}(\mathbf{x}_t^{\mathrm{LR}}, \mathbf{x}_{t-1}^{\mathrm{LR}}).
\end{equation}
The flow field $\mathbf{F}_t$ is also bilinearly up-sampled $\mathcal{S}_{\mathrm{i}\uparrow}^2$ to match the spatial resolution of features in the FAs.
As there is a Deformable Convolutional Layer (DCN) embedded within the FA, we pass the feature responsible for the generation of DCN parameters, i.e\ offsets and masks, across FAs through the DCN Propagation Pathway, acting as residual connection, also known as \textit{Flow Propagation},
\begin{align}
    \{\textcolor{lowresfeat}{\mathbf{h}_t^{l+1}}, \textcolor{dcnprop}{\mathcal{D}_t^{l+1}} \} &= \textcolor{fa}{\mathrm{FA}^l}\left(\mathcal{S}_{\mathrm{s}\downarrow}^4(\textcolor{lowresfeat}{\mathbf{h}_t^l}); \cdot, \cdot, \cdot, \textcolor{dcnprop}{\mathcal{D}_t^l}, \cdot \right), \nonumber \\
    &\qquad  l=0,...,L-1,\\
    \textcolor{dcnprop}{\mathcal{D}_t^0} &= \mathbf{0}.
\end{align}
All FAs except the FA after the up-sampling stage shares the same input configuration.
For the final FA, the Low-Res Feature $\mathbf{h}_{t}^{L-2}$ and DCN parameters $\mathcal{D}_t^{L-2}$ are independently $4\times$ up-sampled using pixel shuffle and convolution.
The previously $2\times$ bilinearly up-sampled flow field is further $4\times$ bilinearly up-sampled.
The fovea-fused feature $\tilde{\mathbf{h}}_{t-1}$ and $\hat{\mathbf{h}}_{t-1}$ bypasses the down-sampler $\mathcal{S}_{\mathrm{s}\downarrow}^4$ in the earlier layers and are fed directly into the final FA.

\begin{figure}[t]
    \centering
    \includegraphics[width=\linewidth]{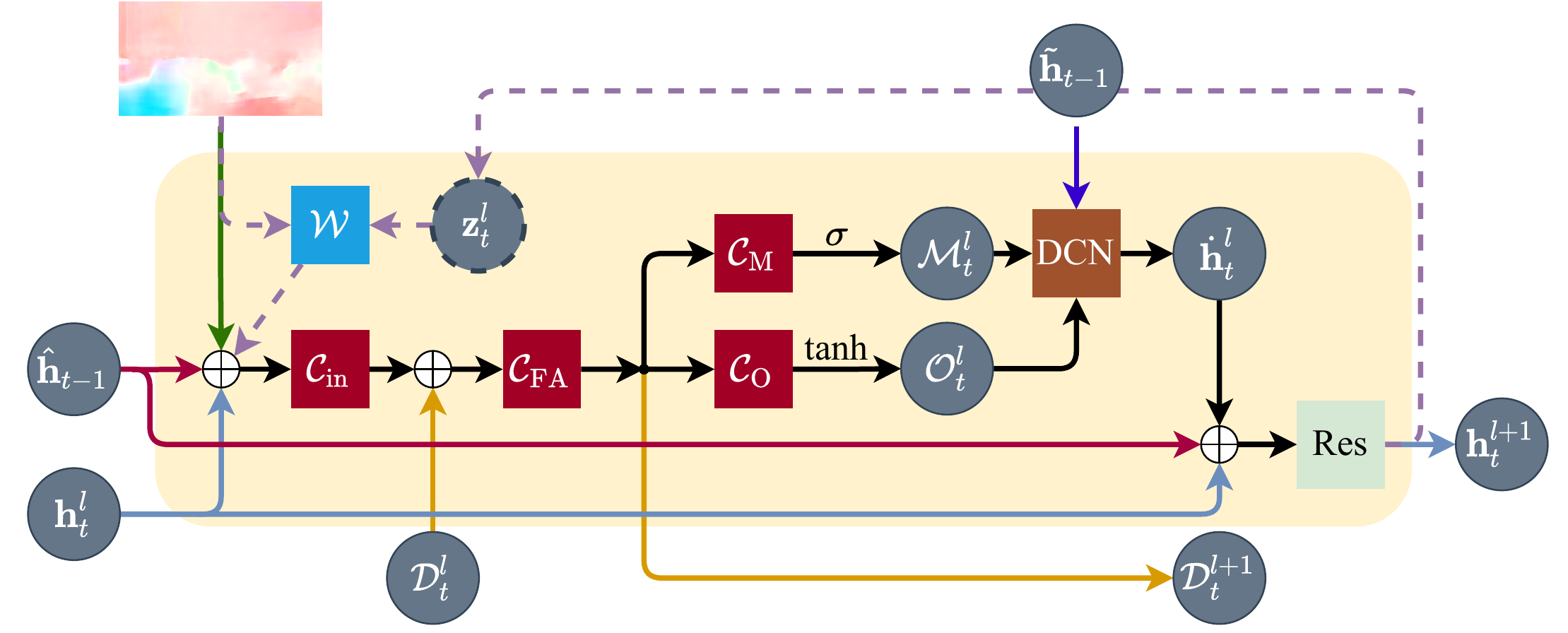}
    \caption{Illustration of \textcolor{fa}{Feature Aggregator (FA)}. $\mathcal{C}$'s are convolutional layers, $\mathrm{DCN}$ is the Deformable Convolutional Layer and $\mathrm{Res}$ is the residual block. DCN warps the fovea-aggregated feature $\tilde{\mathbf{h}}_{t-1}$ from the \textcolor{feedback}{Feedback Pathway} using the estimated offsets $\mathcal{O}_t^l$ and is weighted by estimated masks $\mathcal{M}_t^l$. The result of DCN is concatenated with features from the \textcolor{warpedfeat}{Warped Feature Pathway} $\hat{\mathbf{h}}_{t-1}$ and the \textcolor{lowresfeat}{Low-Res Feature Pathway} $\mathbf{h}_t^l$ and fed to a Residual Block to predict the Low-Res Feature of the upcoming stage $\mathbf{h}_t^{l+1}$ and the \textit{optional} \textcolor{dsv}{DCN state vector (DSV)} $\mathbf{z}_{t+1}^l$. Each FA has its own DSV that involves in the estimation of masks and offsets of DCN. Variables and connections that involves the DSV are represented with \textcolor{dsv}{dashed outline/line}. }
    \label{fig:fa}
\end{figure}

\subsection{Feature Aggregator}\label{sec:aggregator}
With the high-level connections between modules defined, we discuss the inner workings of FA here.
An illustration of FA is shown in Figure \ref{fig:fa}.
In FA, DCN state vector (DSV) $\mathbf{z}_t^l$ along with features from the Warped Feature Pathway $\hat{\mathbf{h}}_{t-1}$, Low-Res Feature Pathway $\mathbf{h}_t^l$ and Flow Field Pathway $\mathbf{F}_t$ are responsible for the estimation of masks $\mathcal{M}_t^l$ and offsets $\mathcal{O}_t^l$ required for DCN,
\begin{align}
    \textcolor{dcnprop}{\mathcal{D}_t^{l+1}} &= \mathcal{C}_{\mathrm{FA}}^l\left(\mathcal{C}_{\mathrm{in}}^l\left(\textcolor{warpedfeat}{\hat{\mathbf{h}}_{t-1}} \oplus \textcolor{lowresfeat}{\mathbf{h}_t^l}  \oplus \right. \right. \nonumber \\
    &\left. \left.  \mathcal{S}_{\mathrm{i}\uparrow}^u(\textcolor{flowfield}{\mathbf{F}_t)}\oplus \mathcal{W}(\textcolor{dsv}{\mathbf{z}_t^l}; \mathcal{S}_{\mathrm{i}\uparrow}^u(\textcolor{flowfield}{\mathbf{F}_t})) \right) \oplus \textcolor{dcnprop}{\mathcal{D}_t^l} \right),\\
    \mathcal{M}_t^l &= \sigma \left( \mathcal{C}^l_{\mathrm{M}}(\textcolor{dcnprop}{\mathcal{D}_t^{l+1}}) \right),\\
    \mathcal{O}_t^l &= \mathrm{tanh}\left( \mathcal{C}^l_{\mathrm{O}}(\textcolor{dcnprop}{\mathcal{D}_t^{l+1}})\vphantom{\mathcal{C}_{\mathrm{FA}}^l}\right).
\end{align}
Note that $u=8$ if $l=L-2$ and $u=2$ otherwise, $\mathcal{C}^l$ are convolutional blocks and $\mathcal{D}_t^l$ is the feature responsible for the estimation of $\mathcal{M}_t^l$ and $\mathcal{O}_t^l$.
$\mathcal{D}_t^l$ is passed to the upcoming DCN block as residual connection.
We can then perform DCN on the feature from the Feedback Pathway $\tilde{\mathbf{h}}_{t-1}$ (down-sampled with $\mathcal{S}_{\mathrm{s}\downarrow} ^4$ for $l<L-2$) as follows,
\begin{align}
    \dot{\mathbf{h}}_{t}^l &= \mathrm{DCN}(\textcolor{feedback}{\tilde{\mathbf{h}}_{t-1}} ; \mathcal{M}^l_{t}, \mathcal{O}^l_{t}),\label{eq:dcn} \\
    &= \mathcal{C}^l_{\mathrm{DCN}}\left( \mathcal{M}^l_{t}\odot\mathcal{W}(\textcolor{feedback}{\tilde{\mathbf{h}}_{t-1}} ; \mathcal{O}^l_{t}) \right),\\
    \dot{\mathbf{h}}_{t}^l (p)&= \sum_{k=1}^K \mathbf{w}_k^l \cdot \textcolor{feedback}{\tilde{\mathbf{h}}_{t-1}} \left(p + p_k + \mathcal{O}^l_{t}(p)\right) \cdot \mathcal{M}^l_{t}(p).
\end{align}
Finally, we can estimate DSV for the next time-step $\mathbf{z}_{t+1}^l$ along with the low-res feature for the upcoming stage $\mathbf{h}_t^{l+1}$ by passing the DCN-warped feature $\dot{\mathbf{h}}_t^l$ along with features from the Warped Feature Pathway $\hat{\mathbf{h}}_{t-1}$ and Low-Res Feature Pathway $\mathbf{h}_t^l$ to a Residual Block,
\begin{equation}
    \{\textcolor{lowresfeat}{\mathbf{h}_t^{l+1}}, \textcolor{dsv}{\mathbf{z}_{t+1}^l} \} = \mathrm{Res}\left(\dot{\mathbf{h}}_{t}^l \oplus \textcolor{warpedfeat}{\hat{\mathbf{h}}_{t-1}} \oplus \textcolor{lowresfeat}{\mathbf{h}_t^l} \right).
\end{equation}
DSV is helpful in retaining information that might potentially be corrupted by the warping operation of the upcoming time-step, e.g.\ future occlusion.

\begin{figure}
    \centering
    \includegraphics[width=\linewidth]{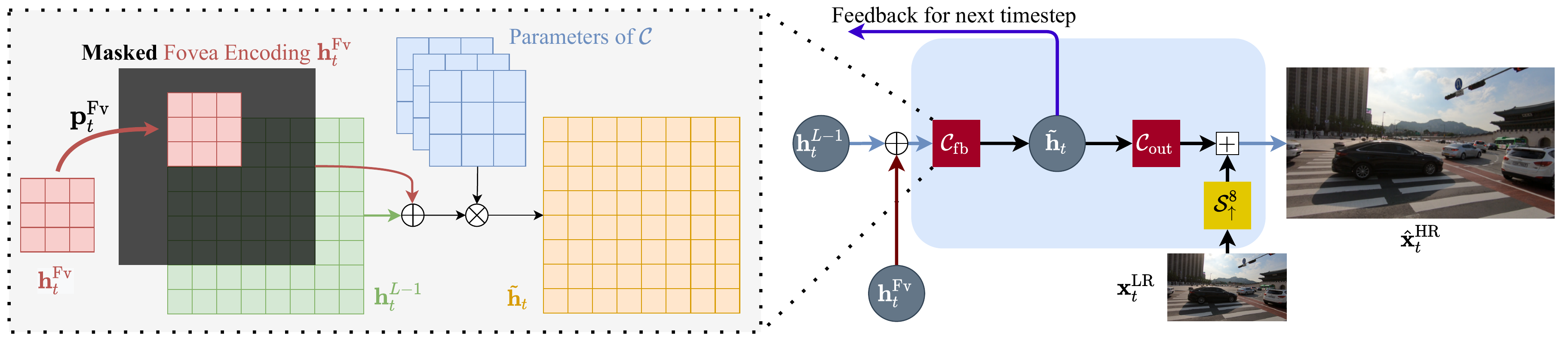}
    \caption{Illustration of \textcolor{ob}{Output Block (OB)}. Given the gaze coordinate $\mathbf{p}_t^{\mathrm{Fv}}$, the encoded foveated region features $\mathbf{h}_t^{\mathrm{Fv}}$ are aligned to the designated position where regional convolution is applied on the concatenated features to generate the fovea-fused feature $\tilde{\mathbf{h}}_t$. $\tilde{\mathbf{h}}_t$ is passed through the \textcolor{feedback}{Feedback Pathway} to provide context from the foveated region to the \textcolor{fa}{FA} blocks. Finally, the frame is rendered at the targeted resolution by adding the estimated result as residual to the bilinearly up-sampled LR frame $\mathcal{S}_{\uparrow}^8(\mathbf{x}_t^{\mathrm{LR}})$.}
    \label{fig:output}
\end{figure}

\subsection{Output Block}\label{sec:output}
After passing through several stages of FAs, comes the final stage where the frame at the targeted resolution will be rendered using the Output Block (OB).
We show an illustration of OB in Figure \ref{fig:output}
The OB takes in $\mathbf{h}_t^{\mathrm{L-1}}$ from the Low-Res Feature Pathway and $\mathbf{h}_t^{\mathrm{Fv}}$ from the Fovea Pathway along with the bilineally up-sampled LR frame $\mathcal{S}_{\uparrow}^8(\mathbf{x}_t^{\mathrm{LR}})$ for frame rendering,
\begin{align}
    \textcolor{feedback}{\tilde{\mathbf{h}}_t} &= \mathcal{C}_{\mathrm{fb}}\left(\textcolor{lowresfeat}{\mathbf{h}_t^{L-1}} \oplus \textcolor{fovea}{\mathbf{h}_t^{\mathrm{Fv}}} \right), \\
    \hat{\mathbf{x}}_t^{\mathrm{HR}} &= \mathcal{C}_{\mathrm{out}}\left( \textcolor{feedback}{\tilde{\mathbf{h}}_t} \right) + \mathcal{S}_{\uparrow}^8\left(\mathbf{x}_t^{\mathrm{LR}}\right).
\end{align}
The estimated results acts are residuals that enhance the bilinearly up-sampled frame.
$\tilde{\mathbf{h}}_t$ contains context from the foveated region and is responsible for the enhancement of feature from earlier stages of the upcoming time-steps.
The quality of $\tilde{\mathbf{h}}_t$ affects the capability of a model to retain HR context that will potentially be picked-up and utilized by the FAs of earlier stages.
As prediction of gaze coordinates using eye trackers usually comes with noise, the capability of a model to retain past HR context is beneficial for the task of FVSR, affecting the overall visual fidelity of transmitted foveated video stream.

\section{Experiments}
This work studies the novel task of FVSR.
Experimental setup is referred from the task of VSR, with several tweaks to shift the focus of experiments towards FVSR.
Since FVSR is targeted for video streaming applications on AR/VR devices that are paired with an eye tracker, we design our experiments to fit such use case through the demonstration of the retention of past foveated region.


\begin{figure*}
    \centering
    \includegraphics[width=\textwidth]{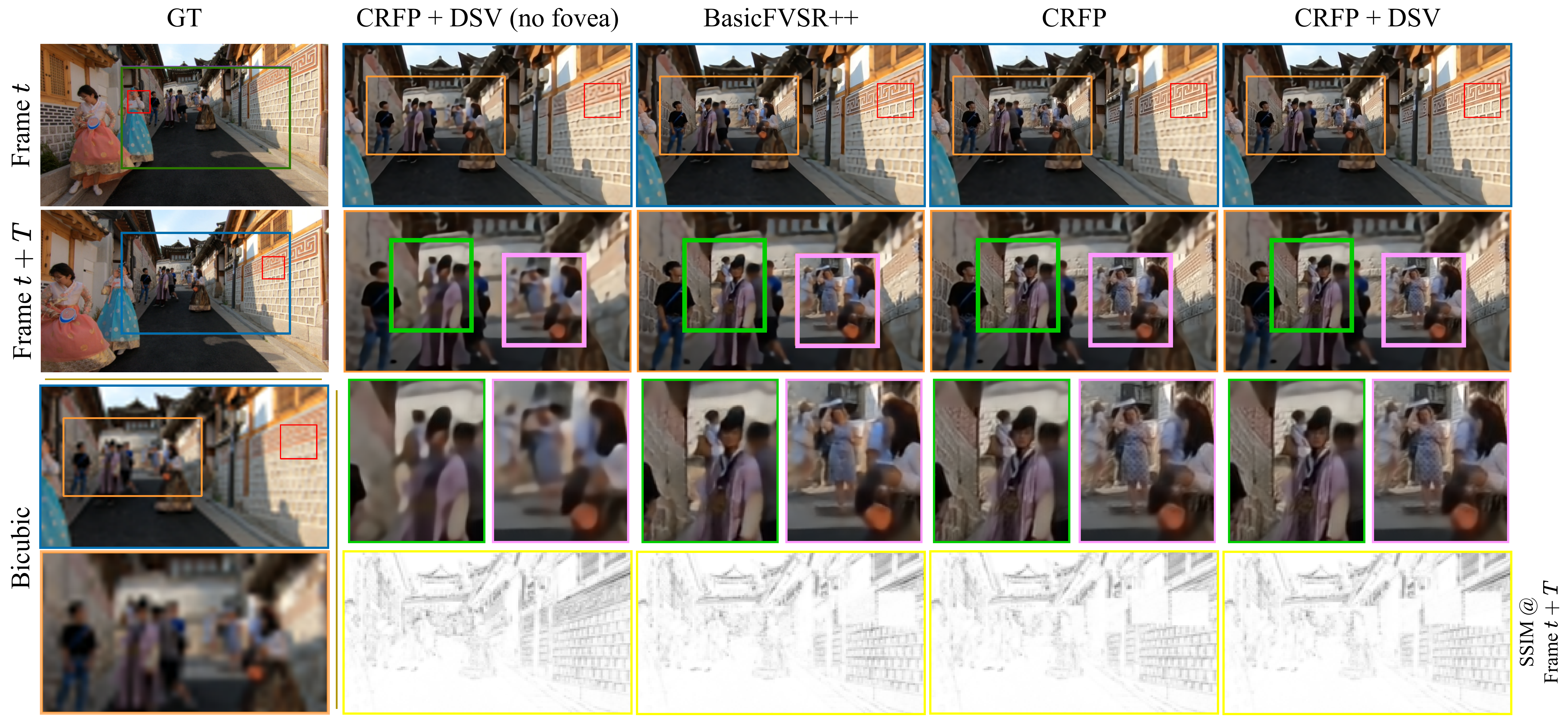}
    \caption{Visual comparison on the task of $8\times$ FVSR. The foveated region (red box) slides from left to right from frame $t$ to frame $t+T$ where $T=6$. PSNR and SSIM plots are for performance comparison, with brighter pixel value indicating higher performance. }
    \label{fig:comparison_general}
\end{figure*}

\paragraph{Dataset.}
Our experiment setup for FVSR bootstraps on the VSR experiments of BasicVSR \cite{chan2021basicvsr,chan2021basicvsr++}.
We train and evaluate on REDS \cite{nah2019ntire}.
We use REDS4\footnote{Clips 000, 011, 015, 020 of REDS training set.} as our test set and REDSval4\footnote{Clips 000, 001, 006, 017 of REDS validation set.} as our validation set.
The remaining clips are used as our training set.
We carry out our study on $8\times$ FVSR since less visual attention is allocated to the region beyond the foveated region.
We apply $8\times$ Bicubic Downsampling to obtain the LR frame for FVSR.

\paragraph{Architecture.}
We use an encoder-decoder for the optical flow estimator $\mathcal{F}$ that is separately trained on MPI Sintel dataset \cite{Butler:ECCV:2012}.
Each convolution block is composed of a single convolutional layer paired with leaky ReLU as its non-linear activation function.
Each up/down-sampling block is composed of a single convolutional layer paired with pixel shuffle operation.
The encoders $\mathcal{E}^{\mathrm{LR}}$ and $\mathcal{E}^{\mathrm{Fv}}$ are composed of 2 convolutional blocks.
For comparison purpose, we modify BasicVSR++ \cite{chan2021basicvsr++} for the task of FVSR.
BasicVSR++ is made causal and foveated region is fed directly to the layer of lowest spatial resolution.
For computationally efficiency, we allocate three FA blocks to the up-sampled encoded LR frame and a single FA block right before the output block.
Detailed configuration is deferred to the supplementary materials.

\paragraph{Training and evaluation.}
We train our models using PyTorch \cite{paszke2017automatic} as our deep learning framework using a single RTX3090 GPU. Runtime in \ref{tab:fvsr_comparison} are measured using frame of size 1080p.
The initial learning rate of our model and flow estimator are set to $1\times 10^{-4}$ and $2.5\times 10^{-5}$ respectively.
The total number of training iterations is 300K with a batch size of 8. 
We use Charbonnier loss \cite{charbonnier1994two} as our loss function for better robustness against outliers.
Images of RED4 are of size $1280\times 720$.
We crop regions of size $256 \times 256$ which are down-sampled to $32\times 32$ as our LR frame during training.
We also crop regions of size $128\times 128$ from the $256\times 256$ patch to be our foveated region.
For training, the coordinate of the foveated region is randomly sampled across whole image.
The coordinate of the foveated region is constrained to not move out of boundary.
For evaluation, we show results that slides foveated region in a raster scan order.
We also show results that has the fovea coordinate oscillate in the vicinity of additive Gaussian noise present in eye trackers to demonstrate the actual use case of CRFP for FVSR task.
The foveated region is cropped to represent $\sim$1\% of the total pixels in a HR image.
We down-sample the LR frames to be of size $160\times 90$ and the cropped foveated region size is $96\times 96$.
We use PSNR, SSIM and VMAF \cite{li2016toward} as metrics to measure the quality of the super-resolved frame.
VMAF is targeted towards video stream and is shown to have higher correlation to the human perception of visual fidelity of video when compared to PSNR and SSIM.

\subsection{In-Depth Study of Foveated Video Super-Resolution}


To evaluate the effectiveness of FVSR, we need to measure the capability of method in retaining HR context of the foveated region and propagating it to future frames.
The fusion of both HR and LR context is also important for optimal visual fidelity.
We can evaluate FVSR using these two regions:
\begin{enumerate}
    \item Foveated region: measures the efficiency in the fusion and the transferring of HR context to the current LR frame.
    \item Past foveated region(s): measures the efficiency in the retaining of HR context of past foveated frame(s) and propagating it to future frames.
\end{enumerate}
As there is no prior work in this field, a fair comparison would be adopting a modification of the SoTA in VSR, BasicVSR++ \cite{chan2021basicvsr++}, to fit the task of FVSR.
We name the modified version as BasicFVSR++.
We modify it to be causal, i.e.\ only frames prior to the current frames are considered for FVSR.
We also reduce its size for the ease of experimentation.
Foveated region is fused with features of earlier layers which differs from our contribution that considers the fusion on a higher spatial resolution.
There are several variations of CRFP that are ablated and studied.
CR\sout{FP} (removal of flow propagation) corresponds to the removal of the DCN Propagation Pathway.
CRFP correspond to the vanilla version that doesn't include the DSV.
CRFP + DSV (no fovea) corresponds to the study of CRFP that includes DSV but without the Fovea Pathway.
CRFP + DSV correspond to the inclusion of the optional DSV.
CRFP-Fast includes DSV and applies the DCN blocks within FA only to a fixed region of size $720\times 720$ to focus on low-latency FVSR.
Quantitative analysis of all regions are summarized in Table \ref{tab:fvsr_comparison}.
From the results, we can see that CRFP is better than BasicFVSR++.
We show the runtime and parameter count of different models for FVSR in Table \ref{tab:fvsr_runtime_param}.
In Figure \ref{fig:comparison_general}, we also show a qualitative comparison of all methods on two different time-steps with a frame interval of $T=6$ in between.
CRFP is visually better when compared to BasicFVSR++.
From the SSIM plots, we can clearly observe past HR contexts are better retained.

\paragraph{Foveated Region.}
We do not directly place the HR frame (foveated region) onto the super-resoluted frame to prevent sharp transition of quality across regions of different resolutions. With the HR frame propagated through a series of convolutional layers, there will definitely be a slight drop in quality as convolutional filtering is a noisy process.
To measure the efficiency in the propagation of fovea information in the main branch, we use PSNR and SSIM as metrics to evaluate the pixels that fall in the foveated region.
Our results show that CRFP outperforms BasicFVSR++ by 4.02 dB. DSV module is also beneficial for the propagation of fovea information across the network with less parameters, showing a marginal boosts of 0.13 dB.
Qualitative results in Figure \ref{fig:comparison_general} show that finer details can be reconstructed by CRFP when compared with BasicFVSR++.

\begin{table*}[]
    \centering
    \caption{Performance comparison of $8\times$ FVSR evaluated using REDS4 at proposed regions using PSNR, SSIM and VMAF.}
    \begin{tabular}{lccccccc}
    \toprule
    \multirow{2}{*}{Method}  & \multicolumn{2}{c}{Foveated Region}  & \multicolumn{2}{c}{Past Foveated Region(s)}   & \multicolumn{3}{c}{Whole Image}  \\
    & PSNR & SSIM & PSNR & SSIM & PSNR & SSIM & VMAF \\
    \midrule
    Bicubic   & 26.16 & 0.6077  & 24.72 & 0.5994  & 23.34 & 0.6077 & 5.1621\\
    BasicFVSR++ \cite{chan2021basicvsr++}  & 37.99 & 0.9560  & 30.36 & 0.8269  & 25.95 & 0.7250 & 68.40\\
    CRFP + DSV (no fovea)   & 29.23 & 0.7128  & 28.24 & 0.7187  & 25.52 & 0.7001 & 64.38\\
    CR\sout{FP}   & 42.07 & 0.9831  & 30.22 & 0.8337  & 25.84 & 0.7202 & 66.32\\
    CRFP   & 42.01 & 0.9835  & \textbf{30.61} & 0.8451  & \textbf{26.13} & 0.7336 & 70.24\\
    CRFP-Fast   & 42.14 & 0.9831  & 29.44 & 0.7983  & 23.72 & 0.6365 & 24.40\\
    CRFP + DSV   & \textbf{42.14}& \textbf{0.9836}  & 30.59 &\textbf{0.8455}  & 26.07 &\textbf{0.7338} &\textbf{70.30}    \\
    \bottomrule
    \end{tabular}
    \label{tab:fvsr_comparison}
    
\end{table*}

\begin{table}[]
    \centering
    \caption{Runtime and model parameters comparison for 1080p video using Nvidia RTX 3090.}
    \begin{tabular}{lcccc}
    \toprule
    Method  & Runtime (ms)  & \# Parameters \\
    \midrule
    BasicFVSR++ \cite{chan2021basicvsr++}  & 35 & 2.35M\\
    CRFP + DSV (no fovea)   & 41 & 2.17M\\
    CR\sout{FP}   & 39 & 2.16M\\
    CRFP   & 42 & 2.21M\\
    CRFP-Fast   & 14 & 2.17M\\
    CRFP + DSV   & 41 & 2.17M\\
    \bottomrule
    \end{tabular}
    \label{tab:fvsr_runtime_param}
    
\end{table}


\begin{figure}
    \centering
    \includegraphics[width=\linewidth]{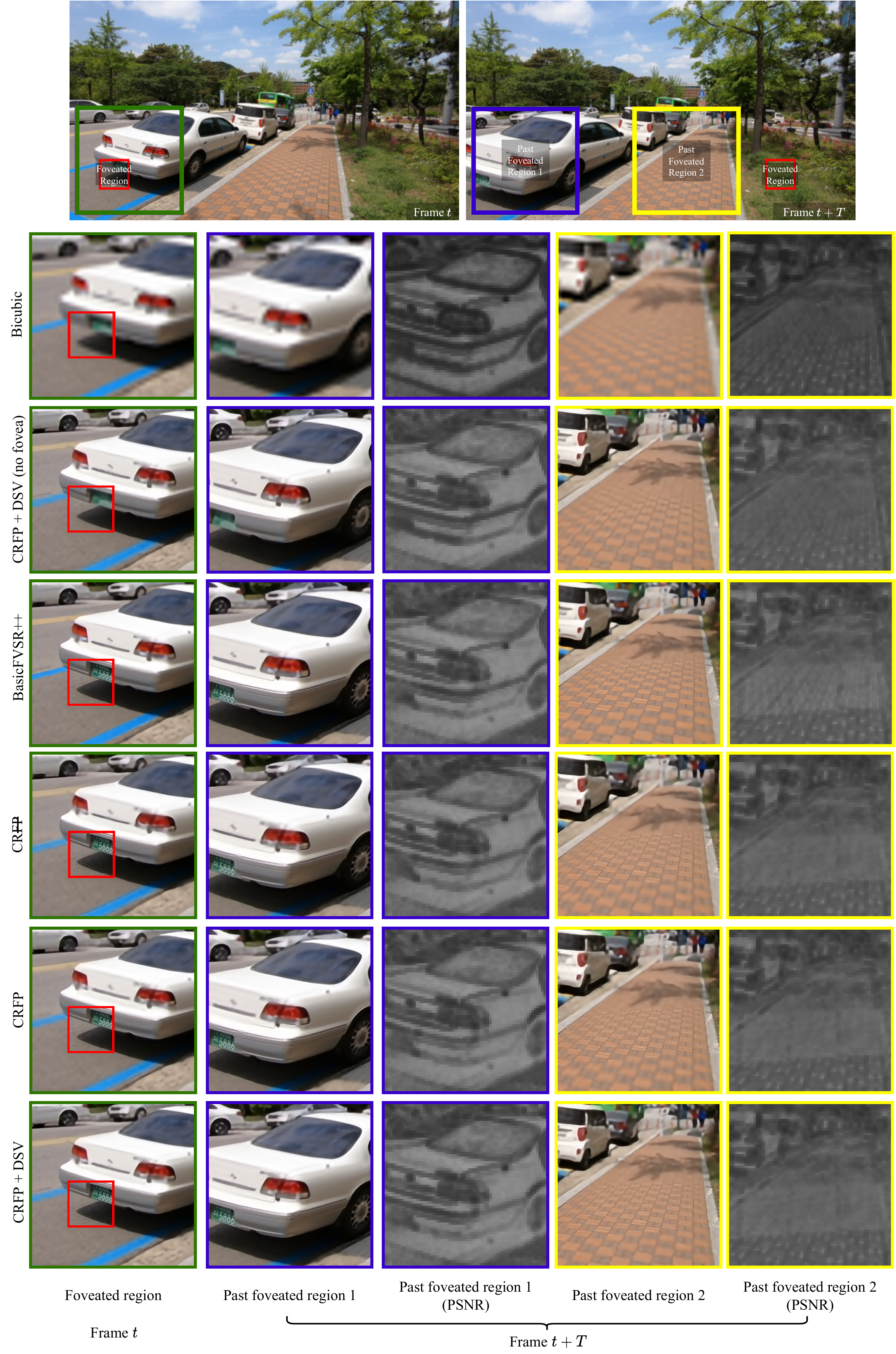}
    \caption{Comparison visual quality of past foveated regions across different methods. Two regions containing high frequency contents are shown. Refer to the license plate of the car and the tiles on the pavement to spot the distinction across different methods.}
    \label{fig:comparison_past}
\end{figure}

\paragraph{Past Foveated Regions.}
An important property of FVSR is the capability of a model to retain HR context(s) that corresponds to foveated region of previous time-step(s). To measure this property, we slide the foveated region across frames using a horizontal trajectory, i.e.\ a straight line from left to right.
The region covered by the trajectory of the foveated region should retain past HR context with high probability.
We evaluate the retention capability of a FVSR model by measuring the PSNR and SSIM at the region covered by the fovea trajectory.
Our results show that CRFP is much better than BasicFVSR++ at retaining information from previous frames, showing 0.25 dB increase in PSNR. 
The inclusion of the DSV module has similar performance as the vanilla variant while requiring less parameters.
The effect of past frame retention is more prominent in the visualizations shown in Figure \ref{fig:comparison_past}. We can see that fine-details from previous HR contexts are still present after an interval of a $T=9$ frames.


\subsection{Simulating FVSR with Eye Tracker Noise}
\begin{figure}
    \centering
    \includegraphics[width=\linewidth]{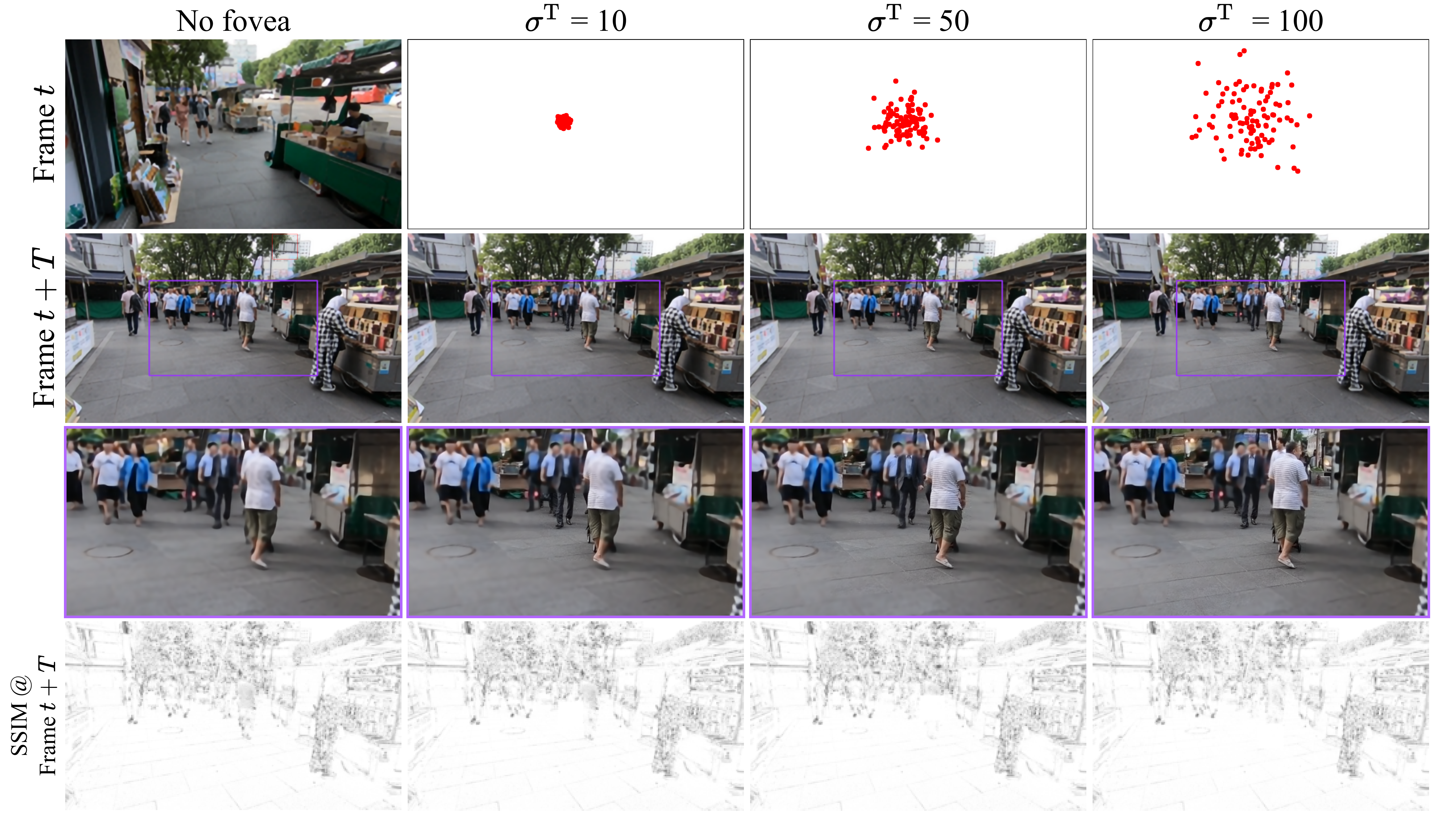}
    \caption{Simulating the actual use case of FVSR where there is additive Gaussian noise present in an eye tracker. Various standard deviations $\sigma^{\mathrm{T}}$ are tested and results show that larger $\sigma^{\mathrm{T}}$ demonstrates the capability of the model on retaining HR context from past foveated regions. Spot the difference on the stripes on the man's shirt and the lines on the ground. SSIM plots are also provided to assist the reader in spotting the differences across different $\sigma^{\mathrm{T}}$.}
    \label{fig:tracker}
\end{figure}

To simulate the application of CRFP to an actual use case of FVSR, we follow the pattern found in eye trackers to influence the trajectory of the foveated region's coordinates.
In Figure \ref{fig:tracker}, we show results for coordinates oscillating under an additive Gaussian noise of $\sigma^\mathrm{T}=10$, $\sigma^\mathrm{T}=50$ and $\sigma^\mathrm{T}=100$.
We can observe that with $\sigma^\mathrm{T}=100$ a larger region can be super-resolved with the context transferred from the past foveated regions.
This experiment shows that the capability of a model on retaining context from past foveated region is a good measure of performance and transfers well to the task of FVSR.
The design of this experiment is to demonstrate the importance of the transferring of context from past foveated region to future frames for the task of FVSR.
The better the performance in retaining context from previous frames the more resilient it is to the noise present in trackers.
The results with $\sigma^\mathrm{T}=100$ demonstrates that context surrounding the gaze region can be clearly reconstructed despite high variation in the eye tracker's reading.

\section{Conclusion}
We propose a novel research direction of Foveated Video Super-Resolution (FVSR) with reliable metrics for the measurement of the foveated visual quality.
The measurement of quality of past foveated region is shown to be beneficial for the task of FVSR through experiments that simulates eye tracker noise.
We show that with induced additive Gaussian noise, CRFP is able to super-resolve context that falls within the specified region through the adoption of context from previous HR foveated region.
Under these metrics, we demonstrate that CRFP is able to perform well on the task of FVSR.
CRFP is designed specifically for FVSR and is suitable for video streaming in AR/VR applications.
CRFP is also designed to be of low-latency, suitable for head-mounted AR/VR devices with limited computational capacity.

\section*{Acknowledgement}
This project is supported by MOST under code MOST 110-2221-E-A49-144-MY3. Eugene Lee is partially supported
by Novatek Ph.D. Fellowship Award. The authors are grate-
ful for the suggestions provided by Dr. Eugene Wong from
University of California in Berkeley and Dr. Jian-Ming Ho
from Academia Sinica of Taiwan.


\clearpage

\appendix

\section{Network Configuration}
All convolutional layers before the final feature aggregator have output channel size of 32, this includes the convolutional layers embedded within the encoder $\mathcal{E}^{\mathrm{LR}}$ and within the feature aggregator blocks.
All convolutional layers are paired with a LeakyReLU activation function to model non-linearity.
The final feature aggregator and the fovea encoder $\mathcal{E}^{\mathrm{Fv}}$ have output channel of size 4.
$\mathcal{C}_{\mathrm{fb}}$ has input channel of 8 (concatenation of foveated region features and features from feature aggregator) and output channel of 4.
$\mathcal{C}_{\mathrm{out}}$ has an output channel of 3.

\paragraph{Flow Field Estimator.}
The flow field estimator $\mathcal{F}$ has an encoder-decoder structure that maps images of the current and previous time step, i.e.\ $\mathbf{I}_t^{\mathrm{LR}}$ and $\mathbf{I}_{t-1}^{\mathrm{LR}}$, to the flow field $\mathbf{F}_t$.
To meet real-time inference latency, we construct our own flow field estimator.
The flow field estimator is composed of 3 encoder blocks, 3 decoder blocks and a flow estimation block.
Both encoder and decoder blocks are composed of two convolutional layers followed by ReLU activation.
Average pooling of kernel size 2 is placed right after each encoding block.
The flow estimation block has two convolutional layer with a ReLU activation layer in between and a tanh activation layer at its output.

\section{Experiments on DCN State Vector}
DCN state vectors (DSV) are introduced to retain state information that are useful in super-resolving future frames.
The introduction of DSV helps in reducing the required parameters and computational cost as less features are propagated towards the upcoming feature aggregators and are stored internally as state vectors within the feature aggregator blocks.
Here, we perform a study on the trade-off between the allocation of features for forward propagation or are propagated internally within each feature aggregators as DSV.
We summarize the ablation study on DSV in Table \ref{tab:dsv-ablate}.
We can observe that the introduction of a small amount of DSV into the feature aggregator contributes to the final performance.

\begin{table*}[]
    \centering
\scriptsize
    \caption{Performance comparison of $8\times$ FVSR evaluated using REDS4 at proposed regions using PSNR, SSIM and VMAF. Comparison using various input configuration of the feature aggregation is shown. Setting the total input channels as 32, we study the trade-off in ratio between the features from the previous feature aggregator ($\hat{\mathbf{h}}_{t-1}\oplus \mathbf{h}_t^l$) and the DSV embedded within the current feature aggregator.}
    \begin{tabular}{ccccccccc}
    \toprule
    \multicolumn{2}{c}{\parbox{2.5cm}{\centering Channels}}   & \multicolumn{2}{c}{\parbox{3cm}{\centering Foveated Region}}  & \multicolumn{2}{c}{\parbox{3.5cm}{\centering Past Foveated Region(s)}}   & \multicolumn{3}{c}{\parbox{2.5cm}{\centering Whole Image}}  \\
    $\hat{\mathbf{h}}_{t-1}\oplus \mathbf{h}_t^l$  & DSV &  PSNR & SSIM & PSNR & SSIM & PSNR & SSIM & VMAF \\
    \midrule
     8 & 24  & \textbf{42.27} & 0.9835  & 30.29 & 0.8361  & 25.87 & 0.7246 & 67.12\\
     16 & 16 & 42.12 & 0.9834  & 30.47 & 0.8424  & 25.96 & 0.7292 & 69.88\\
     24 & 8  & 42.14 & \textbf{0.9836}  & \textbf{30.59} &\textbf{0.8455}  & \textbf{26.07} &\textbf{0.7338} &\textbf{70.30}    \\
     32 & 0  & 41.31 & 0.9831 & 29.96 & 0.8242 & 25.78 & 0.7182 & 66.58    \\
    \bottomrule
    \end{tabular}
    \label{tab:dsv-ablate}
    
\end{table*}

\section{Simulating FVSR with Eye Tracker Noise}
\begin{figure*}
    \centering
    \includegraphics[width=\linewidth]{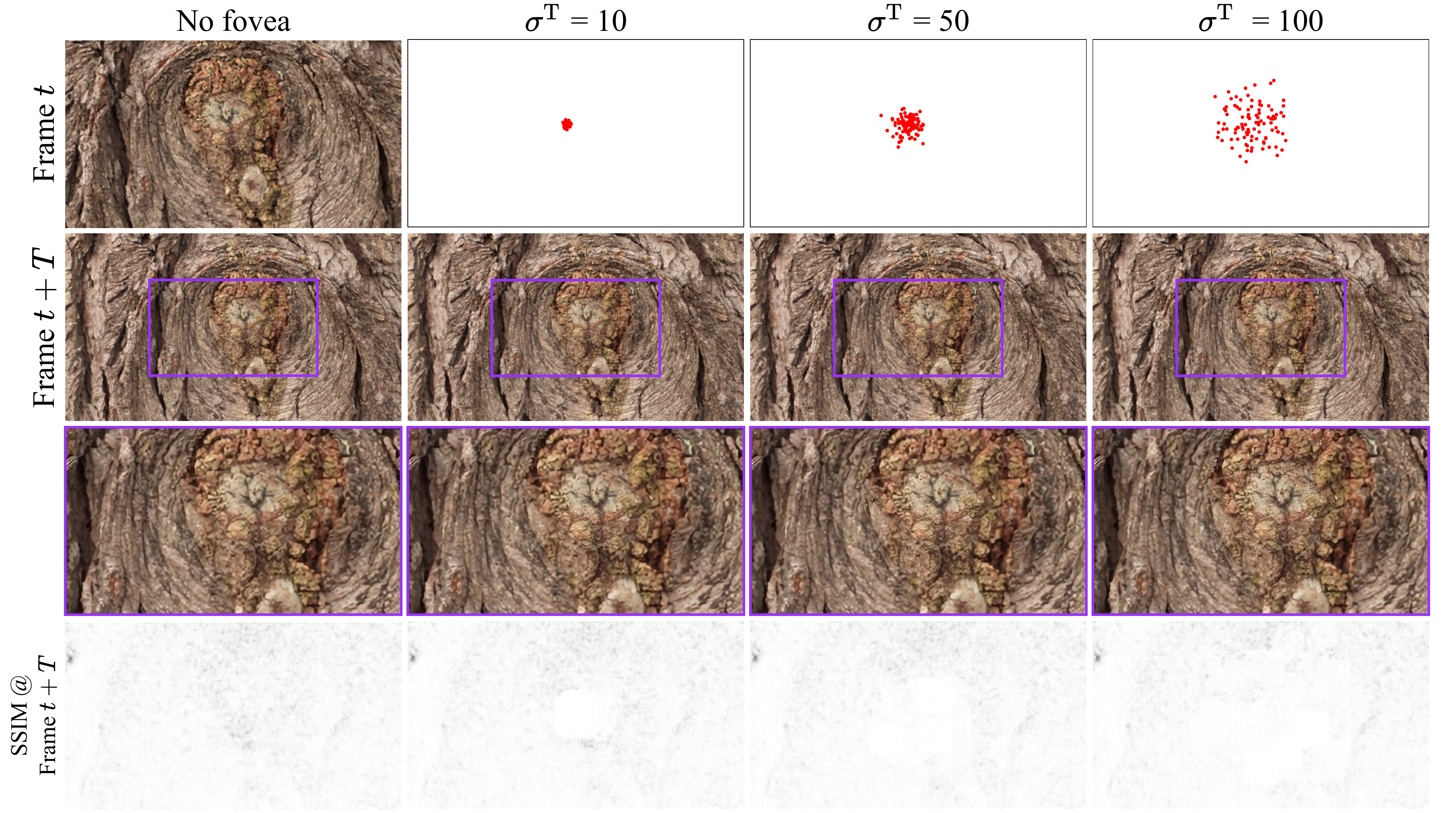}
    \caption{Simulating the actual use case of FVSR where there is additive Gaussian noise present in an eye tracker. Various standard deviations $\sigma^{\mathrm{T}}$ are tested and results show that larger $\sigma^{\mathrm{T}}$ demonstrates the capability of the model on retaining HR context from past foveated regions. Spot the difference in details of the stripes on the log. SSIM plots are also provided to assist the reader in spotting the differences across different $\sigma^{\mathrm{T}}$. Larger $\sigma^{\mathrm{T}}$ results in larger coverage of HR region but loses marginal detail at the center point.}
    \label{fig:eyetracker}
\end{figure*}

We show similar a simulation as the one shown in the main paper in Figure \ref{fig:eyetracker}.

\section{Analysis of CRFP-Fast}
\begin{figure*}
    \centering
    \includegraphics[width=\linewidth]{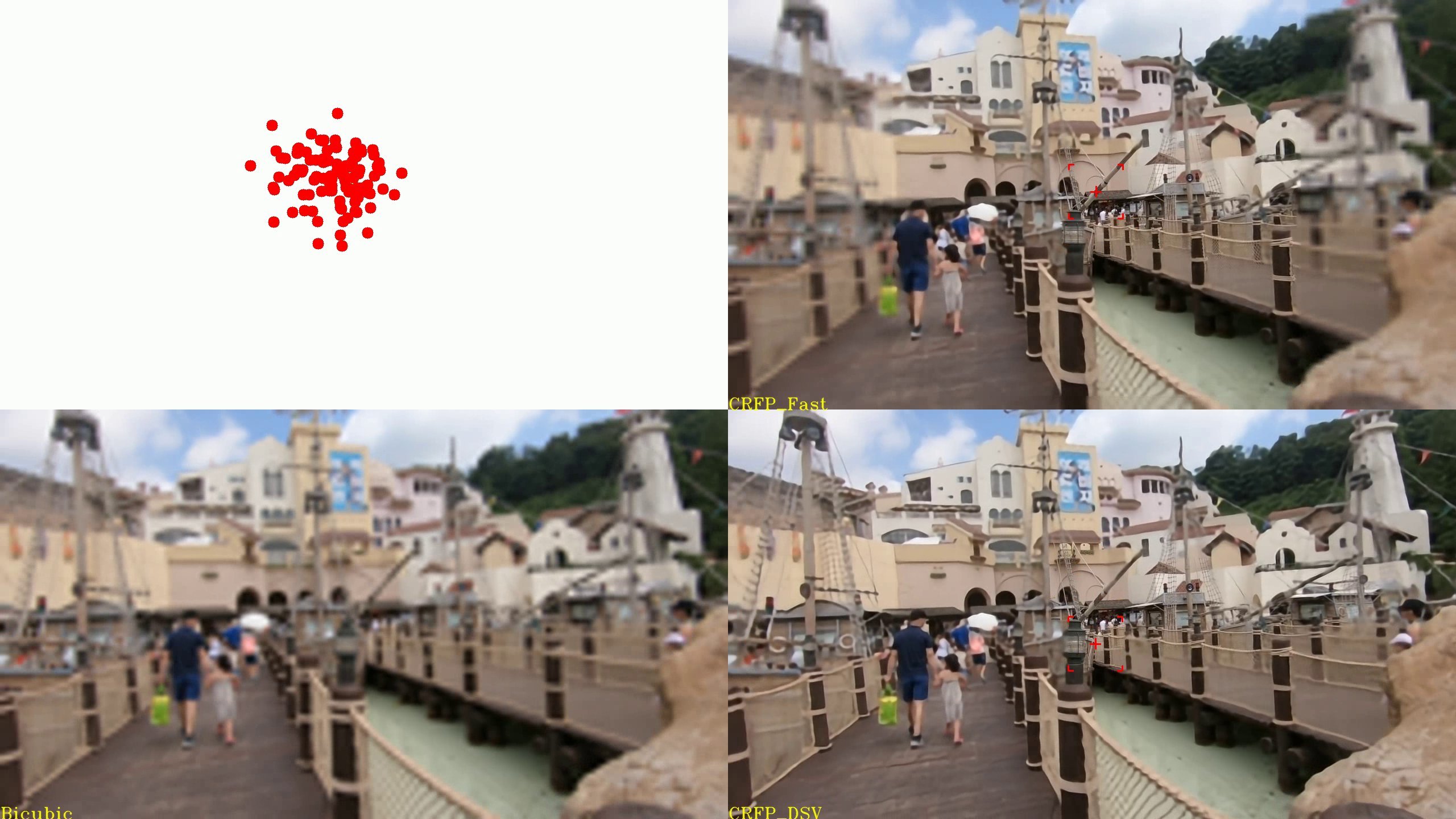}
    \caption{\textbf{Top Left}: 100 center points of foveated region; \textbf{Top Right}: CRFP-Fast after 100 frames; \textbf{Bottom Right}: CRFP + DSV after 100 frames; \textbf{Bottom Left}: Bicubic result after 100 frames. Notice that while CRFP has noticeably lower quality beyond the region ($720\times 720$) passed into the DCN, it is not visually perceptible if we focus on the foveated region.}
    \label{fig:crfp-fast}
\end{figure*}

For CRFP to achieve real-time latency for head-mounted device, only a fixed region ($720\times 720$) is passed through the DCN blocks within the feature aggregator for fine-grained warping while the rest are forward propagated through the residual block within the feature aggregator.
Using this approach, we are able to reduce the latency by a factor of 3 (latency of 14 ms per frame using RTX 3090), enabling real-time inference using our architecture.
Although CRFP-Fast has low VMAF score in the main paper, it is shown to be visually pleasing in Figure \ref{fig:crfp-fast}.
As pixel region far beyond the foveal acuity are not efficiently picked up by our visual system, loss in visual quality in that region doesn't not affect our visual perception of the video.

\section{Video in Supplementary Materials}
We show videos with the format of Figure \ref{fig:crfp-fast} in our supplementary materials.
The name formatting of the videos follows the rule \texttt{$\sigma^{\mathrm{T}}$\_videoID.mp4}.
$\sigma^{\mathrm{T}}$ correspond to the standard deviation of the distribution of the additive Gaussian noise introduced to the foveated region.

{\small
\bibliographystyle{ieee_fullname}
\bibliography{egbib}
}

\end{document}